\documentclass{article}
\usepackage{spconf,amsmath,graphicx,bm}

\usepackage{url}
\usepackage{multirow}
\usepackage[inline]{enumitem}
\usepackage{amsfonts}

\def\x{{\mathbf x}}
\def\X{{\mathbf X}}
\def\z{{\mathbf z}}
\def\Z{{\mathbf Z}}
\def\bmu{{\bm \mu}}
\def\tmu{{\bm {\tilde \mu}}}
\def\L{{\cal L}}

\title{Extracting Domain Invariant Features by Unsupervised Learning \\for Robust Automatic Speech Recognition}
\name{Wei-Ning Hsu, James Glass}
\address{MIT Computer Science and Artificial Intelligence Laboratory \\
         Cambridge, MA 02139, USA\\
         \texttt{\{wnhsu,glass\}@mit.edu}}

\begin{document}
\ninept

\maketitle

\begin{abstract}
The performance of automatic speech recognition (ASR) systems can be significantly compromised by previously unseen conditions, which is typically due to a mismatch between training and testing distributions.
In this paper, we address robustness by studying domain invariant features, such that domain information becomes transparent to ASR systems, resolving the mismatch problem. 
Specifically, we investigate a recent model, called the Factorized Hierarchical Variational Autoencoder (FHVAE).
FHVAEs learn to factorize sequence-level and segment-level attributes into different latent variables without supervision.
We argue that the set of latent variables that contain segment-level information is our desired domain invariant feature for ASR.
Experiments are conducted on Aurora-4 and CHiME-4, which demonstrate 41\% and 27\% absolute word error rate reductions respectively on mismatched domains.
\end{abstract}
\begin{keywords}
robust speech recognition, factorized hierarchical variational autoencoder, domain invariant representations
\end{keywords}

\section{Introduction}
Recently, neural network-based acoustic models~\cite{sainath2015deep,sak2015acoustic,hsu2016prioritized} have greatly improved the performance of automatic speech recognition (ASR) systems. 
Unfortunately, it is well known (e.g., ~\cite{hsu2017unsuperviseddomain}) that ASR performance can degrade significantly when testing in a domain that is mismatched from training.
A major reason is that speech data have complex distributions and contain information about not only linguistic content, but also speaker identity, background noise, room characteristics, etc. 
Among these sources of variability, only a subset are relevant to ASR, while the rest can be considered as a nuisance and therefore hurt the performance if the distributions of these attributes are mismatched between training and testing. 

To alleviate this issue, some robust ASR research focuses on mapping the out-of-domain data to in-domain data using enhancement-based methods~\cite{narayanan2013ideal,isik2016single,feng2014speech}, which generally requires parallel data from both domains.
Another popular strategy is to train an ASR system with as large, and as diverse a dataset as possible~\cite{li2012improving,seltzer2013investigation};
however, this strategy is not feasible when the labeled data are not available for all domains.
Alternatively, robustness can also be achieved by training using features that are domain invariant~\cite{kingsbury1998robust,stern2012features,vinyals2011comparing,sainath2012auto,sun2017unsupervised}. 
In this case, we would not have domain mismatch issues, because domain information is now transparent to the ASR system.

In this paper, we consider the same highly adverse scenario as in~\cite{hsu2017unsuperviseddomain}, where both clean and noisy speech are available, but the transcripts are only available for clean speech.
We study the use of a recently proposed model, called Factorized Hierarchical Variational Autoencoder (FHVAE)~\cite{hsu2017unsupervised}, for learning domain invariant ASR features without supervision. 
FHVAE models learn to factorize sequence-level attributes and segment-level attributes into different latent variables. 
By training an ASR system on the latent variables that encode segment-level attributes, and testing the ASR in mismatched domains, we demonstrate that these latent variables contain linguistic information and are more domain invariant.
Comprehensive experiments study the effect of different FHVAE architectures, training strategies, and the use of derived domain features on the robustness of ASR systems.
Our proposed method is evaluated on Aurora-4~\cite{pearce2002aurora} and CHiME-4~\cite{vincent2016analysis} datasets, which contain artificially corrupted noisy speech and real noisy speech respectively.
The proposed FHVAE-based feature reduces the absolute word error rate (WER) by 27\% to 41\% compared to filter bank features, and by 14\% to 16\% compared to variational autoencoder-based features.
We have released the code of FHVAEs described in the paper.\footnote{\url{https://github.com/wnhsu/FactorizedHierarchicalVAE}}

The rest of the paper is organized as follows. 
In Section~\ref{sec:method}, we introduce the FHVAE model and a method to extract domain invariant features. Section~\ref{sec:setup} describes the experimental setup, while Section~\ref{sec:exp} presents results and discussion.
We conclude our work in Section~\ref{sec:conclu}.

\section{Learning Domain Invariant Features}\label{sec:method}

\subsection{Modeling a Generative Process of Speech Segments}
As mentioned above, generation of speech data often involves many independent factors, which are however unseen in the unsupervised setting.
It is therefore natural to describe such a generative process using a latent variable model, where a latent variable $\z$ is first sampled from a prior distribution, and a speech segment $\x$ is then sampled from a distribution conditioned on $\z$. 
In~\cite{hsu2017learning}, a convolutional variational autoencoder (VAE) is proposed to model such process; 
by assuming the prior to be a diagonal Gaussian, it is shown that the VAE automatically learns to model independent attributes regarding generation, such as the speaker identity and the linguistic content, using orthogonal latent subspaces.
This result provided a mechanism of potentially learning domain invariant features for ASR by discovering latent variables that do not contain domain information.

\subsection{Extracting Domain Invariant Features from FHVAEs}
The generation of sequential data often involves multiple independent factors operating at different scales.
For instance, the speaker identity affects the fundamental frequency (F0) at the utterance level, while the phonetic content affects spectral characteristics at the segment level.
As a result, sequence-level attributes, such as F0 and volume, tends to have a smaller amount of variation within an utterance, compared to between utterances, while the other attributes, such as spectral contours, tend to have similar amounts of variation within and between utterances.

Based on this observation, FHVAEs~\cite{hsu2017unsupervised} formulate the generative process of sequential data with a factorized hierarchical graphical model that imposes sequence-dependent priors and sequence-independent priors to different sets of latent variables. 
Specifically, given a dataset $\mathcal{D} = \{ \X^{(i)} \}_{i=1}^M$ consisting of $M$ i.i.d. sequences, where $\X^{(i)} = \{ \x^{(i,n)} \}_{n=1}^{N^{(i)}}$ is a sequence of $N^{(i)}$ segments (sub-sequence), a sequence $\X$ of $N$ segments is assumed to be generated from a random process that involves latent variables $\Z_1 = \{ \z_1^{(n)} \}_{n=1}^N$, $\Z_2 = \{ \z_2^{(n)} \}_{n=1}^N$, and $\bmu_2$ as follows:
\begin{enumerate*}[label=(\arabic*)]
	\item an \textit{s-vector} $\bmu_2$ is drawn from a prior distribution $p_{\theta}(\bmu_2) = \mathcal{N}(\bmu_2 | \bm{0}, \sigma^2_{\bmu_2}\bm{I})$;
    \item $N$ i.i.d. \textit{latent segment variables} $\{ \z_1^{(n)} \}_{n=1}^N$ and \textit{latent sequence variables} $\{ \z_2^{(n)} \}_{n=1}^N$ are drawn from a sequence-independent prior $p_{\theta}(\z_1) = \mathcal{N}(\z_1 | \bm{0}, \sigma^2_{\z_1}\bm{I})$ and a sequence-dependent prior $p_{\theta}(\z_2 | \bmu_2) = \mathcal{N}(\z_2 | \bmu_2, \sigma^2_{\z_2}\bm{I})$ respectively;
    \item $N$ i.i.d. speech segments $\{ \x^{(n)} \}_{n=1}^N$ are drawn from a condition distribution $p_{\theta}(\x | \z_1, \z_2) = \mathcal{N}(\bm{x} | f_{\mu_x}(\z_1, \z_2), diag(f_{\sigma^2_x}(\z_1, \z_2)))$, whose mean and diagonal variance are parameterized by neural networks. 
\end{enumerate*}
The joint probability for a sequence is formulated in Eq.~\ref{eq:joint}:
\begin{equation}
	p_{\theta}(\bmu_2) \prod_{n=1}^{N} p_{\theta}(\x^{(n)} | \z_1^{(n)}, \z_2^{(n)}) p_{\theta}(\z_1^{(n)})p_{\theta}(\z_2^{(n)} | \bmu_2). \label{eq:joint}
\end{equation}
Based on this formulation, $\bmu_2$ can be regarded as a summarization of sequence-level attributes for a sequence, 
and $\z_2$ is encouraged to encode sequence-level attributes for a segment that are similar within an utterance.
Consequently, $\z_1$ encodes the residual segment-level attributes for a segment,
such that $\z_1$ and $\z_2$ together provide sufficient information for generating a segment.

Since exact posterior inference is intractable, FHVAEs introduce an inference model $q_{\phi}( \Z_1^{(i)}, \Z_2^{(i)}, \bmu_2^{(i)} | \bm{X}^{(i)})$ as formulated in Eq.~\ref{eq:inf} that approximates the true posterior $p_{\theta}(\Z_1^{(i)}, \Z_2^{(i)}, \bmu_2^{(i)} | \bm{X}^{(i)})$:
\begin{align}
    &q_{\phi}( \bmu_2^{(i)}) \prod_{n=1}^{N^{(i)}} q_{\phi}(\z_1^{(i,n)} | \x^{(i,n)}, \z_2^{(i,n)}) q_{\phi}(\z_2^{(i,n)} | \bm{x}^{(i,n)}), \label{eq:inf}
\end{align}
from which we observe that inference of $\z_1^{(i,n)}$ and $\z_2^{(i,n)}$ only depends on the corresponding segment $\x^{(i,n)}$;
in particular, the posteriors, 
$q_{\phi}(\z_1 | \x, \z_2) = \mathcal{N}(\z_1 | g_{\mu_{\z_1}}(\x, \z_2), diag( g_{\sigma^2_{\z_1}}(\x, \z_2) ) )$ and 
$q_{\phi}(\z_2 | \x) = \mathcal{N}(\z_2 | g_{\mu_{\bm{z}_2}}(\x), diag( g_{\sigma^2_{\z_2}}(\x) ) )$, are approximated with diagonal Gaussian distributions whose mean and diagonal variance are also parameterized by neural networks.
On the other hand, $q_{\phi}(\bmu_2^{(i)})$ is modeled as an isotropic Gaussian, $\mathcal{N}(\bmu_2^{(i)} | g_{\mu_{\bmu_2}}(i), \sigma^2_{\tmu_2}\bm{I})$, where $g_{\mu_{\bmu_2}}(i)$ is a trainable lookup table of the posterior mean of $\bmu_2$ for each training sequence.
Estimation of $\bmu_2$ for testing sequences can be found in~\cite{hsu2017unsupervised}.

As pointed out in~\cite{hsu2017unsuperviseddomain}, nuisance attributes regarding ASR, such as speaker identity, room geometry, and background noise, are generally consistent within an utterance.
If we treat each utterance as a sequence, these attributes then become sequence-level attributes, which would be encoded by $\z_2$ and $\bmu_2$.
As a result, $\z_1$ encodes the residual linguistic information and is invariant to these nuisance attributes, which is our desired domain invariant ASR feature. 

\subsection{Training FHVAE and Preventing S-Vector Collapsing}
As in other generative models, FHVAEs aim to maximize the marginal likelihood of the observed dataset;
due to the intractability of the exact posterior, FHVAEs optimize the \textit{segment variational lower bound}, $\L(\theta, \psi; \x^{(i,n)})$, which is formulated as follows:
\begin{align}
	& \mathbb{E}_{q_{\phi}(\bm{z}_1^{(i,n)}, \bm{z}_2^{(i,n)} | \bm{x}^{(i,n)})} \big[ 
    		\log p_{\theta}(\bm{x}^{(i,n)} | \bm{z}_1^{(i,n)}, \bm{z}_2^{(i,n)}) \big] \nonumber \\
    & - \mathbb{E}_{q_{\phi}(\bm{z}_2^{(i,n)} | \bm{x}^{(i,n)})} \big[ 
    		D_{KL}(
            	q_{\phi}(\bm{z}_1^{(i,n)} | \bm{x}^{(i,n)}, \bm{z}_2^{(i,n)}) || 
            	p_{\theta}(\bm{z}_1^{(i,n)})) \big] \nonumber \\
    & - D_{KL}( 
        	q_{\phi}(\bm{z}_2^{(i,n)} | \bm{x}^{(i,n)}) || 
            p_{\theta}(\bm{z}_2^{(n)} | g_{\mu_{\bmu_2}}(i))) + \dfrac{1}{N} \log p_{\theta}(g_{\mu_{\bmu_2}}(i)). \nonumber
\end{align}
Notice that if the $\bmu_2$ are the same for all utterances, an FHVAE would then degenerate to a vanilla VAE.
To prevent $\bmu_2$ from collapsing, we can add an additional discriminative objective, $\log p(i|\z_2^{(i,n)})$, that encourages the discriminability of $\z_2$ regarding which utterance the segment is drawn from.
Specifically,  we define it as 
$\log p_\theta( \z_2^{(i,n)} | g_{\mu_{\bmu_2}}(i)) - \log \sum_{j=1}^M p_\theta( \z_2^{(i,n)} | g_{\mu_{\bmu_2}}(j) ) $.
By combining the two objectives with a weighting parameter $\alpha$, we obtain the \textit{discriminative segment variational lower bound}: 
\begin{equation}
	\mathcal{L}^{dis}(\theta,\phi; \bm{x}^{(i,n)})
    	= \mathcal{L}(\theta,\phi; \bm{x}^{(i,n)})
        + \alpha \log p(i | \bm{z}_2^{(i,n)}). \label{eq:lb_dis}
\end{equation}

\begin{table*}[t!]
  \centering
  \begin{tabular}{|llllll|l|llll|}
    \hline
    \multicolumn{6}{|c}{Setting}	& \multicolumn{1}{|c|}{WER (\%)} & \multicolumn{4}{c|}{WER (\%) by Condition} \\
    Exp. Index	& Feature 	& \#Layers	& \#Units	& $\alpha$	& Seq. Label 	& Avg.	& A	& B	& C	& D	\\
    \hline
    \multirow{4}{*}{1}	& FBank		& -	& -	& - & - 			& 65.64	& 3.21	& 61.61	& 51.78	& 82.39 \\
                & $\z$		& 1/1	& 256/256	& -		& -		& 44.79	& 4.22	& 38.16	& 36.11	& 59.63 \\
                & $\z$		& 1/1 	& 512/256	& -		& -		& 40.31	& 4.35	& 33.83	& 34.43	& 53.77 \\
                & $\z_1$	& 1/1 	& 256/256	& 10	& uttid	& \textbf{26.58} & 4.54	& 19.28	& 20.85	& 38.50\\
    \hline\hline
    \multirow{3}{*}{2}	& $\z_1$	& 1/1 	& 256/256	& 10	& uttid	& 26.58 & 4.54	& 19.28	& 20.85	& 38.50\\
                & $\z_1$	& 2/2 	& 256/256	& 10	& uttid	& 25.54 & 4.11	& 16.90	& 20.62	& 38.58 \\
                & $\z_1$	& 3/3 	& 256/256	& 10	& uttid	& \textbf{24.30} & 4.91	& 15.44	& 22.83	& 36.63 \\
    \hline\hline
    \multirow{3}{*}{3}	& $\z_1$	& 1/1 	& 128/128	& 10	& uttid	& 34.66 & 5.06	& 26.70	& 25.39	& 49.09 \\
                & $\z_1$	& 1/1 	& 256/256	& 10	& uttid	& \textbf{26.58} & 4.54	& 19.28	& 20.85	& 38.50\\
                & $\z_1$	& 1/1 	& 512/512	& 10	& uttid	& 26.97 & 5.32	& 18.18	& 23.13	& 40.01	\\
    \hline\hline
    \multirow{5}{*}{4}	& $\z_1$	& 1/1 	& 256/256	& 0		& uttid	& 33.30 & 4.86	& 25.67	& 25.46	& 46.97 \\
                & $\z_1$	& 1/1 	& 256/256	& 5		& uttid	& 30.55 & 4.63	& 22.66	& 23.33	& 43.96	\\
                & $\z_1$	& 1/1 	& 256/256	& 10	& uttid	& \textbf{26.58} & 4.54	& 19.28	& 20.85	& 38.50 \\
                & $\z_1$	& 1/1 	& 256/256	& 15	& uttid	& 29.92 & 5.01	& 20.82	& 24.79	& 44.03 \\
                & $\z_1$	& 1/1 	& 256/256	& 20	& uttid	& 32.64 & 5.57	& 25.48	& 24.53	& 45.66 \\
    \hline\hline
    \multirow{3}{*}{5}	& $\z_1$	& 1/1 	& 256/256	& 10	& uttid	& \textbf{26.58} & 4.54	& 19.28	& 20.85	& 38.50\\
                & $\z_1$	& 1/1 	& 256/256	& 10	& noise	& 32.27 & 4.33	& 23.89	& 28.96	& 45.86 \\
                & $\z_1$	& 1/1 	& 256/256	& 10	& speaker	& 34.95 & 4.39	& 27.27	& 32.22	& 48.20 \\
    \hline\hline
    \multirow{2}{*}{6}	& $\z_1$	& 1/1 	& 256/256	& 10	& uttid	& \textbf{26.58} & 4.54	& 19.28	& 20.85	& 38.50\\
                & $\z_1$-$\bmu_2$	& 1/1 & 256/256	& 10	& uttid	& 43.61	& 5.08	& 42.47	& 27.55	& 53.85 \\
    \hline
  \end{tabular}
  \caption{Aurora-4 test\_eval92 set word error rate of acoustic models trained on different features.}
  \label{tab:a4_wer}
\end{table*}

\section{Experiment Setup}\label{sec:setup}
To evaluate the effectiveness of the proposed method on extracting domain invariant features, we consider domain mismatched ASR scenarios. Specifically, we train an ASR system using a clean set, and test the system on both a clean and noisy set. The idea is that one would observe a smaller performance discrepancy between different domains if the feature representation is more domain invariant. We next introduce the datasets, as well as the model architectures and training configurations for the experiments.

\subsection{Dataset}
We use Aurora-4~\cite{pearce2002aurora} as the primary dataset for our experiments. Aurora-4 is a broadband corpus designed for noisy speech recognition tasks based on the Wall Street Journal (WSJ0) corpus~\cite{garofalo2007csr}. 
Two microphone types, clean/channel are included, and six noise types are artificially added to both microphone types, which results in four conditions: clean(A), channel(B), noisy(C), and channel+noisy(D). 
We use the multi-condition development set for training the VAE and FHVAE models, because the development set contains both noise labels and speaker labels for each utterance, which are used in \textit{Exp. Index 5}, while the training set only contains speaker labels. 
The ASR system is trained on the clean \texttt{train\_si84\_clean} set and evaluated on the multi-condition \texttt{test\_eval92} set.

To verify our proposed method on a non-artificial dataset, we repeat our experiments on the CHiME-4~\cite{vincent2016analysis} dataset, which contains real distant-talking recordings in noisy environments.
We use the original 7,138 clean utterances and the 1,600 single channel real noisy utterances in the training partition to train the VAE and FHVAE models.
The ASR system is trained on the original clean training set and evaluated on the CHiME-4 development set.

\subsection{VAE/FHVAE Setup and Training}
The VAE is trained with stochastic gradient descent using a mini-batch size of 128 without clipping to minimize the negative variational lower bound plus an $L2$-regularization with weight $10^{-4}$. The Adam~\cite{kingma2014adam} optimizer is used with $\beta_1=0.95$, $\beta_2=0.999$, $\epsilon=10^{-8}$, and initial learning rate of $10^{-3}$. Training is terminated if the lower bound on the development set does not improve for 50 epochs.
The FHVAE is trained with the same configuration and optimization method, except that the loss function is replaced with the negative discriminative segment variational lower bound.

Seq2Seq-VAE~\cite{hsu2017unsuperviseddomain} and Seq2Seq-FHVAE~\cite{hsu2017unsupervised} architectures with LSTM units are used for all experiments.
We let the latent space of the VAEs contain 64 dimensions.  Since the FHVAE models have two latent spaces, we let each of them be 32 dimensional.
Other hyper-parameters are explored in our experiments.
Inputs to VAE/FHVAE, $\x$, are chunks of 20 consecutive speech frames randomly drawn from utterances, where each frame is represented as 80 dimensional filter bank (FBank) energies.
To extract features from the VAE and FHVAE for ASR training, for each utterance, we compute and concatenate the posterior mean and variance of chunks shifted by one frame, which generates a sequence of new features that are 19 frames shorter than the original sequence.
We pad the first frame and the last frame at each end to match the original length.

\subsection{ASR Setup and Training}
Kaldi~\cite{povey2011kaldi} is used for feature extraction, decoding, forced alignment, and training of an initial HMM-GMM model on the original clean utterances. 
The recipe provided by the CHiME-4 challenge (\texttt{run\_gmm.sh}) and the Kaldi Aurora-4 recipe are adapted by only changing the training data being used.
The Computational Network Toolkit (CNTK)~\cite{yu2014introduction} is used for neural network-based acoustic model training.
For all experiments, the same LSTM acoustic model~\cite{sak2014long} with the architecture proposed in~\cite{zhang2016highway} is applied, which has 1,024 memory cells and a 512-node projection layer for each LSTM layer, and 3 LSTM layers in total.

Following the training setup in~\cite{hsu2016exploiting}, LSTM acoustic models are trained with a cross-entropy criterion, using truncated backpropagation-through-time (BPTT)~\cite{williams1990efficient} to optimize. 
Each BPTT segment contains 20 frames, and each mini-batch contains 80 utterances, since we find empirically that 80 utterances has similar performance to 40 utterances. 
A momentum of 0.9 is used starting from the second epoch~\cite{hsu2016prioritized}. 
Ten percent of the training data is held out as a validation set to control the learning rate. The learning rate is halved when no gain is observed after an epoch. The same language model is used for decoding for all experiments.

\begin{table*}[tbh]
  \centering
  \begin{tabular}{|llllll|cc|cccc|}
    \hline
    \multicolumn{6}{|c}{Setting}	& \multicolumn{2}{|c|}{WER (\%)} & \multicolumn{4}{c|}{WER (\%) by Noise Type} \\
    Exp. Index	& ASR Feature 	& \#Layers	& \#Units	& $\alpha$	& Seq. Label 	& Clean	& Noisy	& BUS	& CAF	& PED	& STR	\\
    \hline
	\multirow{3}{*}{1}	& FBank		& -		& -			& -		& -		& 19.37 & 87.69	& 95.56	& 92.05	& 78.77	& 84.37	\\
                & $\z$		& 1/1 	& 512/256	& -		& -		& 19.47	& 73.95	& 70.10	& 91.45	& 64.26	& 69.99	\\ 
                & $\z_1$	& 1/1 	& 256/256	& 10	& uttid	& 19.57	& \textbf{67.94}	& 71.96	& 79.37	& 59.32	& 61.11	\\    
   	\hline\hline
    \multirow{3}{*}{2}	& $\z_1$	& 1/1 	& 256/256	& 10	& uttid	& 19.57	& 67.94	& 71.96	& 79.37	& 59.32	& 61.11	\\
                & $\z_1$	& 2/2 	& 256/256	& 10	& uttid	& 19.73	& 62.44	& 71.28	& 71.86	& 52.46	& 54.18	\\
                & $\z_1$	& 3/3 	& 256/256	& 10	& uttid	& 19.52	& \textbf{60.39}	& 69.13	& 66.24	& 51.22	& 54.96	\\
    \hline
  \end{tabular}
  \caption{CHiME-4 development set word error rate of acoustic models trained on different features.}
  \label{tab:chime4_wer}
\end{table*}

\section{Experimental Results and Discussion}\label{sec:exp}
In this section, we report the experimental results on both datasets, and provide insights on the outcome.
Table~\ref{tab:a4_wer} and \ref{tab:chime4_wer} summarize the results on Aurora-4 and CHiME-4 respectively.
For both tables, different experiments are separated by double horizontal lines and indexed by the \textit{Exp. Index} on the first column.
The second column, \textit{Feature}, refers to the frame representations used for training ASR models.
The third to the sixth column explains the model configuration and the discriminative training weight for VAE or FHVAE models.
We separate the encoder and decoder parameters by ``/'' in the third and the fourth column.
Averaged and by-condition word error rate (WER) are shown in the rest of the columns.

\subsection{Baseline}
We start with establishing Aurora-4 baseline results trained on different types of feature representations, including 
\begin{enumerate*}[label=(\arabic*)]
	\item FBank, 
    \item latent variable, $\z$, extracted from the VAE, and 
    \item latent segment variable, $\z_1$, extracted from the FHVAE. 
\end{enumerate*}
Because each FHVAE model has two encoders, to have a fair comparison between VAE and FHVAE models, we also consider a VAE model with 512 hidden units at each encoder layer.
The results are shown in Table~\ref{tab:a4_wer} \textit{Exp. Index 1}.
As mentioned, condition A is the matched domain, while conditions B, C, and D are all mismatched domains.

FBank degrades significantly in the mismatched conditions, producing between 49\% to 79\% absolute WER increase. 
On the other hand, both VAE and FHVAE models improve the performance in the mismatched domains by a large margin, with only a slight degradation in the matched domain. 
In particular, the features learned by the FHVAE consistently outperform the VAE features in all mismatched conditions by 14\% absolute WER reduction.

We believe that this experiment verifies that FHVAEs can successfully retain domain invariant linguistic features in $\z_1$, while encode domain related information into $\z_2$. 
In contrast, as the results suggests, VAEs encode all the information into a single set of latent variables, $\z$, which still contain domain related information that can hurt ASR performance on the mismatched domains.

\subsection{Comparing Model Architectures}
We next explore the optimal FHVAE architectures for extracting domain invariant features.
In particular, we study the effect of the number of hidden units at each layer and the number of layers.
Results of each variant are listed in Table~\ref{tab:a4_wer} \textit{Exp. Index 2} and \textit{Exp. Index 3} respectively.
Regarding the averaged WER, the model with 256 hidden units at each layer and in total three layers achieves the lowest WER (24.30\%).
Interestingly, if we break down the WER by condition, it can be observed that increasing the FHVAE model capacity (i.e. increasing number of layers or hidden units) helps reducing the WER in the noisy condition (B), but deteriorates channel-mismatching condition (C) above 256 hidden units and 2 layers.

\subsection{Effect of FHVAE Discriminative Training}
Speaker verification experiments in~\cite{hsu2017unsupervised} suggest that discriminative training facilitates factorizing segment-level attributes and sequence-level attributes into two sets of latent variables. Here we study the effect of discriminative training on learning robust ASR features, and show the results in Table~\ref{tab:a4_wer} \textit{Exp. Index 4}. When $\alpha = 0$, the model is not trained with the discriminative object. 
While increasing the discriminative weight from 0 to 10, we observe consistent improvement in all 4 conditions due to better factorization of segment and sequence information; however, when further increasing the weight to 20, the performance starts to degrade. This is because the discriminative object can inversely affects the modeling capacity by constraining the expressibility of the latent sequence variables.

\subsection{Choice of Sequence Label}
A core idea of FHVAE is to learn sequence-specific priors to model the generation of sequence-level attributes, which have a smaller amount of variation within a sequence. 
Suppose we treat each utterance as one sequence, then both speaker and noise information belongs to sequence-level attributes, because they are consistent within an utterance. 
Alternatively, we consider two FHVAE models that learn speaker-specific priors and noise-specific priors respectively. 
This can be easily achieved by concatenating sequences of the same speaker label or noise label, and treating it as one sequence used for FHVAE training. 
We report the results in Table~\ref{tab:a4_wer} \textit{Exp. Index 5}.

It may at first seem surprising that utilizing supervised information in this fashion does not improve performance. 
We believe that concatenating utterances actually discards some useful information with respect to learning domain invariant features. 
FHVAEs use latent segment variables to encode attributes that are not consistent within a sequence.
By concatenating speaker utterances, noise information is no longer consistent within sequences, and would thus be encoded into latent segment variables;
similarly, latent segment variables would not be speaker invariant in the other case.

\subsection{Use of S-Vector}
Lastly, we study the use of s-vectors, $\bmu_2$, derived from the FHVAE model, which can be seen as a summarization of sequence-level attributes of an utterance. We apply the same procedure as i-vector based speaker adaptation~\cite{saon2013speaker}: For each utterance, we first estimate its s-vector, and then concatenate s-vectors with the feature representation of each frame to generate the new feature sequence.

Results are shown in Table~\ref{tab:a4_wer} \textit{Exp. Index 6}, from which we observe a significant degradation of WER that is similar to those of the VAE models. 
This is reasonable because $\z_1$ and $\bmu_2$ in combination actually contains similar information as the latent variable $\z$ in VAE models, and the degradation is due to the mismatch between the distributions of $\bmu_2$ in the training and testing sets.

\subsection{Verifying Results on CHiME4}
In this section, we repeat the baseline and the layer experiments on the CHiME-4 dataset, in order to verify the effectiveness of the FHVAE and the optimality of the FHVAE architecture on a non-artificial dataset. 
The results are shown in Table~\ref{tab:chime4_wer}. 
From \textit{Exp. Index 1}, we see that the same trend applies to the CHiME-4 dataset, where the latent segment variables from the FHVAE outperform those from the VAE, and both latent variable representations outperform FBank features.
For the FHVAE architectures, a 7\% absolute WER decrease is achieved by increasing the number of encoder/decoder layers from 1 to 3, which is also consistent with the trends we saw on Aurora-4.

\section{Conclusion and Future Work}\label{sec:conclu}
In this paper, we conduct comprehensive experiments on studying the use of FHVAE models domain invariant ASR features extractors. 
Our feature demonstrates superior robustness in mismatched domains compared to FBank and VAE-based features by achieving 41\% and 27\% absolute WER reduction on Aurora-4 and CHiME-4 respectively.
In the future, we plan to study FHVAE-based augmentation methods similar to~\cite{hsu2017unsuperviseddomain}.

\vfill
\pagebreak
\clearpage

\bibliographystyle{IEEEbib}
\bibliography{refs}

\end{document}